# TOWARDS TEXT-BASED PHISHING DETECTION

**Gilchan Park and Julia M. Taylor**
**Department of Computer and Information Technology**
**Purdue University**
**West Lafayette, IN 47907, USA**

**ABSTRACT**

This paper reports on an experiment into text-based phishing detection using readily available resources and without the use of semantics. The developed algorithm is a modified version of previously published work that works with the same tools. The results obtained in recognizing phishing emails are considerably better than the previously reported work; but the rate of text falsely identified as phishing is slightly worse. It is expected that adding semantic component will reduce the false positive rate while preserving the detection accuracy.

**INTRODUCTION**

Phishing is one of the most potentially disruptive actions that can be performed on the internet. Stealing a user's account information within a business network through a phishing scam can be an easy way to gain access to that business network. Intellectual property and other pertinent business information could potentially be at risk if a user behind the keyboard falls for a phishing attack. The most common way of carrying out a phishing attack is through email. While such attacks may be easily identifiable for those well-versed in technology, it may be difficult for the typical internet user to spot a fraudulent email.

According to RSA (Brook 2012), there have been nearly 33 thousand phishing attacks each month in 2012. The total loss from these attacks as of August of 2012 was around $690 million, and over $1.5 billion (Kessem 2013) by the end of the year. This is a 59% increase from 2011 in the number of phishing attacks and 22% higher in the losses. It clear that the phishing attacks are on the rise. The results of the phishing attack can be devastating, as a recent example of the attack on South Korea showed: an attack that shut down several prominent South Korean banks and broadcasters in March 2013 had an origin as a spear phishing campaign (Donohue 2013).

The recommended defense against phishing attacks is to educate a user not to fall for them. Unfortunately, the awareness campaigns are not the most effective solution: it only takes one employee to fall for the bait for a company and for its clients to become vulnerable. Because phishing emails rely on human emotions, it is essential to provide an emotion-free solution to a phishing detection game. Such a solution should work before a user can click on the fraudulent link and, thus, before any damage can be made. At the same time, if a user is well educated and understands that an email is a phishing attempt, it should be possible for the user to indicate it as such, so that the solution software has a chance to validate its results or correct its evaluation based on real-time data.

To make it more complicated, phishing emails can target a particular individual or a small group of individuals based on their known behaviors. Thus, it is important for the software to adapt to individual characteristics of an individual or a group of individuals and predict their vulnerabilities for phishing.

There is a need for a phishing-detection application to understand "the buttons" that are being pushed by fraudulent emails. When such buttons are detected, it is possible to react to them in a desired way. To date, there have been a number of papers that report detection of phishing emails (Hong 2012, Basnet et al 2008, Cleber et al 2011, Mutton 2011, Aaron 2010, Xiang 2011, Shahriar & Zulkernine 2012) based on non-text features claiming an accuracy over 90%. Yet, the problem still exists and growing. One possible explanation of this is these methods don't address the content of the messages that people fall for.

**BACKGROUND**

Phishing is a malicious use of internet resources done to trick internet users to reveal personal information to the attacker. An attack is typically performed by sending an email to an unsuspecting user that contains a link to a domain that is seemingly legitimate in the hopes that the user will input their private information for the attacker to steal (DigiCert 2009). Phishing is a criminal act which uses a combination of "social engineering and technical subterfuge" to steal user information (APWG 2012). The most common type of phishing attack leverages email. The emphasis of this paper is to detect phishing attempts within emails.

The threat phishing poses to Internet users at large calls for action within the information security industry to create ways of detecting and preventing such attacks. Research into the area of phishing detection has yielded several types of email analysis to determine if an email should be classified as phishing such as link analysis, header analysis, and text analysis. Link analysis refers to the using information about the links included within an email to determine whether the email is legitimate or a phishing attempt. This usually involves checking to see if the displayed link in the email matches the actual website URL that the user is taken to if the link is clicked. Header analysis refers to examining the header contents of an email to decide if the email is a phishing email or not. This analysis typically includes checking that the 'From' field of the email matches the actual sender and checking the IP address from which the email was sent against phishing blacklists. Blacklist is a set of well-known phishing Web sites and addresses reported by trusted entities such as Google's and Microsoft's black list (Gaurav et al. 2012). For black listing, both a client and a server side are necessary. The client component's implementation can be done through an email or browser plug-in that communicate with a server component. The server component is a public Web site containing a list of phishing sites (Tout & Hafner 2009). Text analysis refers to the examination of text included in the body of the email to find out if an email has a suspicious content.

**Text Analysis in Phishing Detection**

We will outline several recent phishing detection systems that use text analysis. One of them is Phish Mail Guard, a Phishing Mail Detection Technique that uses textual and URL analysis and a phishing detection method based on a combination of blacklist, white list and some heuristics (Hajgude & Ragha 2012). DNS analyzer component in the system determines whether email is phishing or non phishing by analyzing visual DNS and actual DNS in email. If the DNS of hyperlink is present in blacklist, email is considered as phishing. If it is present in white list, email is considered as non-phishing. Black list contains list of known fake DNS and the white list holds list of known valid DNS. DNS analyzer module is implemented using those lists to select a technique for further examination. If the DNS of hyperlink in email does not fall into either category, heuristic detection process takes over the next step. The heuristic module has text and URL algorithms. For the text algorithm, body of text in the email is parsed into tokens and the tokens are compared to blacklisted token. If there are more matched tokens with blacklisted tokens than some threshold, the email is considered to be phishing. In the URL algorithm, link URL in body text is parsed into tokens and compared with blacklisted features from URL. For example, the numbers of @ symbol, the length of hostname and IP address in URL are counted. Similarly, if the number of matched tokens is more than some threshold, it is considered to be phishing.

CANTINA (Zhang et al. 2007) and CANTINA+ (Xiang et al. 2011) are content-based approach for detecting phishing web sites, which examine the content of a web page to determine whether it is phishing. CANTINA uses TF-IDF and the Robust Hyperlink algorithms. TF-IDF is often used in information retrieval and text mining and measures how important a word is to a document in a corpus. TF means the number of times a given term appears in a specific document. IDF represents a measure of the general importance of the term. In other words, it shows how common a term is across an entire collection of documents. If a term has a high TF-IDF weight, TF is high and DF is low. Robust Hyperlink algorithm is developed to solve "404 not found" (broken links) problem (Phelps & Wilensky 2000). Lexical signatures are a small number of well chosen terms to identify the given page. Lexical signatures are added to URLs and if the link does not work, then it feeds signatures to search engine. TF-IDF is adapted to generate useful lexical signatures and the researchers found that top five words as scored by TF-IDF were surprisingly effective. CANTINA is based on two assumptions: that scammers often directly copy legitimate web pages or include keywords like name of legitimate organization, and with Google, phishing web pages should have a low Google Page Rank due to a lack of links pointing to the scam. In the CANTINA process, first, it calculates the TF-IDF score for each word in a given web page. Second, it takes five words with highest TF-IDF weights. Third, it feeds those five keywords in the Google search engine. If the domain name of current web page is in top N search results, it considers email legitimate. CANTINA defines N = 30 since the number 30 worked well. As a means of reducing false positive, heuristic methods are applied to CANTINA. First method is to add the domain name to the lexical signature since the domain name itself usually can best identify the web page. Second method is called ZMP. If Google returns zero search results, the web site is regarded as phishing. Even though this technique has the potential to increase false positive, when combined with adding the domain name, it can actually reduce false positive. For the last, CANTINA added several heuristics from SpoofGuard (Chou et al. 2004), and PILFER (Fette et al. 2007) -- the well-known phishing detection tools. They tested 200 URLs linked to English language sites, and compared CANTINA with SpoofGuard and Netcraft anti-phishing toolbars. The results showed that CANTINA had a true positive rate of 89%, SpoofGuard had 91%, and Netcraft had 97%. When it comes to the false positive rate, CANTINA had 1%, SpoofGuard had 48%, and Netcraft had 0%.

PhishNet-NLP (Verma, Shashidhar, & Hossain, 2012) is a phishing detection algorithm based on email contents analysis using natural language techniques. PhishNet-NLP is designed to distinguish between "actionable" and "informational" emails. The main idea of PhishNet-NLP is that phishing emails are designed to trigger an action from users. Therefore, the "actionable"

email is referred to email leading users to do certain actions in email texts. On the other hand, the "informational" email means legitimate emails. PhishNet-NLP makes use of all the information in emails, except attachments: header, links, text. The analyzing email text module is based on natural language techniques including parsing, part of speech tagging, named entity recognition, stemming, stopword removal and word sense disambiguation. The algorithm consists of a combination of link analysis, header analysis, and text analysis, and it determines if the email poses a phishing threat by a total score that is a sum of results of three analyses. The algorithm produced promising results in the study. Header and link analysis in their study consistently performed with an accuracy rating of over 95% in detecting phishing emails in the experiments run. Text analysis lagged behind, performing between about 60% and 80% accuracy. The studies presented in this paper focus on improvement of the accuracy rating of the text analysis that is based on the idea proposed in PhishNet-NLP.

**THE EXPERIMENT**

The result of PhishNet-NLP shows that link and header analysis are superior to text analysis. The purpose of this research was to expand the text analysis portion so as to fill in the gap left by link and header analysis as well as determine whether it is possible to improve PhishNet-NLP's overall performance of email analysis capabilities. The text analysis portion of PhishNet-NLP takes into consideration actionable verbs that tempt the user into performing an action.

In this study, the modified algorithm includes not only actionable verbs, but also other Parts Of Speech (POS) so that it can catch any other actionable words in phishing emails, not just verbs. It is based on an intuition that a command "*Update* your…" can be as easily made with "your account information needs to be *updated*" or "An *update* of your account" where, in this case, *update* is the action in question.

Two corpora will be applied to the experiments for testing a new algorithm. The first corpus used will be the phishing corpus (Nazario 2004) used in the original PhishNet-NLP experiments. The total number of emails contained in this corpus is 4558, all of which are phishing emails. The other corpus used for the experiments is the Enron email corpus (CALO Project 2009). This is also a publicly available email corpus. The chosen collection from Enron corpus for this implementation contained 7944 emails, all of which are legitimate emails.

The Stanford POS tagger version 3.1.4 wsj-0-18-left3words was used for POS tagging. The Stanford Tokenizer was used to divide text into a sequence of words. The stemming algorithm, which is essential to extract the stem of the word in the emails, used in this project was Porter stemmer (Porter 1980). Finally, WordNet (Miller et al. 1990, Fellbaum 2010) 2.1 was used to generate synsets of the words used in analysis. WordNet 2.1 was used to match as closely as possible the setup of the original PhishNet-NLP experiments.

As not all details could be obtained from the original PhishNet-NLP paper or its authors, we had to make the following assumptions. First, the actionable words for use within these experiments were selected based on the set of the *sample* keywords supplied by the authors of PhishNet-NLP within their paper. The authors did not explicitly state what keywords to use within PhishNet-NLP. Second, as the list of stopwords used in the original PhishNet-NLP was not provided, we used the default English stopwords list found at the following location: http://www.ranks.nl/resources/stopwords.html. Third, synonyms and troponyms of actionable words were chosen for the experiments by the first sense of actionable words.

Within the original PhishNet-NLP experiments, SenseLearner (Mihalcea & Csomai, 2005) and TextRank (Mihalcea & Tarau, 2004) were used for word sense disambiguation. These tools were not used in this experiment design due to difference in programming languages used. WordNet orders senses by the estimated usage frequency of each sense of a word (Du et al., 2008). The most frequently used sense of each word was therefore used to find synonyms and troponyms in this analysis. Forth, no context score will be covered in these experiments, unlike the original PhishNet-NLP experiments. This is due to a lack of clarity in how context score evaluation took place within the original algorithm. The context score is the similarity computation between the new email and the emails received in the past. Lastly, in the original PhishNet-NLP, the Named Entity Recognition technique was used to check that a phishing email has a recipient's name. However, since the email recipients' names were not provided, this experiment did not need to utilize the Stanford Named Entity Recognizer.

**RESULTS AND DISCUSSIONS**

The experiments were performed on phishing and non-phishing corpus to check the accuracy of prediction/classification. As we were not able to replicate the original PhishNet-NLP experiments, there is some difference in the results within the original PhishNet-NLP results and our implementation of that algorithm. The original results are shown in Figure 1.

**Phishing Corpus**

Testing on the phishing corpus yielded similar (although not the same) results for the original PhishNet-NLP algorithm with context score removed that was achieved in the original experiments. The original PhishNet-NLP algorithm in this simulation was able to correctly identify 65.6% of phishing emails within the phishing corpus. Our modified algorithm showed an identification of 83.5% of phishing email within the phishing corpus, which is an 18% increase in identification over the PhishNet-NLP algorithm, identifying. The exact numbers are shown in Figure 2.

First reason that contributes to improvement in phishing detection was that actionable keywords were found not only as verb forms, but also other POS (see Figure 3). The Stanford POS abbreviations can be found in Figure 6. The new expanded algorithm found 40% actionable keywords of all keywords that it founded from NN (noun), VBG (gerund) and VBN (past participle). The sum of those forms is larger than the portion of VB (verb). Even assuming that all forms of the verb were accounted for (VBN, VBG, etc), the sum of their tags with VB still only accounts for 70% of the data contributing to classification. This result shows that in many cases actionable keywords can exist in different POS forms, and it means that it must consider other POS when an actionable keyword is added.

It is likely that Stanford parser's errors contribute to the better results when all parts of speech are used: for NN (noun) and NNP (singular pronoun), Stanford parser often tags a verb as NN or NNP in case that a sentence is an imperative sentence. For example, in the sentence, "Click here to verify your account if you choose to ignore our request", the verb "Click" is tagged as NNP; in the sentence, "Please visit PayPal as soon as possible to verify your identity" the verb visit is tagged as NN (noun). The expanded algorithm covers all POS, and that lead to significant improvement in phishing detection.

**Legitimate Corpus**

Testing on the Enron corpus yielded slightly better results for the original PhishNet-NLP algorithm over this extended algorithm. The original algorithm was able to correctly label 91.2% of the emails in the Enron corpus as legitimate. The expanded algorithm was able to correctly label 85.1% of the emails as legitimate. This means that the new algorithm had about a 6% increase in false positive rates for the Enron corpus. The results of this analysis can be seen in Figure 4.

Since the new algorithm considers all POS, it catches and calculates more keywords than the original algorithm does. That means the expanded algorithm could mistake legitimate emails for phishing emails. In the result of the expanded algorithm shown in the figure below, VBG (gerund), NNP (singular pronoun) and NN (noun) had significantly impact on incensement in false positive rates. The portion of gerund, noun and pronoun is 44% of all keywords that lead false positive. Even though those POS forms improve to detect phishing emails, at the same time they play a key role in increasing false positive rates.

The proposed algorithm resulted in the increase in false positive rate. Even though 6% increase may not be substantial, false positive in emails should cause worse problems to users. It could be argued the risk to the organization from one missed phishing email is larger than from one legitimate email flagged as dangerous. Nevertheless, it is still a problem that should be addressed.

To address this increased false positive issue, possible solutions are suggested. First, Assigning different values to words in the set of special verbs used in the algorithm should increase accuracy of the algorithm. The frequent words used by the algorithm were different between the legitimate and phishing corpora. Word frequencies are presented in the figures below. The most frequent words in phishing emails were click, update, use, and confirm. The most frequent words in the legitimate corpus were subject, go, see, and click. In the current algorithm, all words in the set of special verbs used have the same 'L' value in the equation: score (v) = $\{1 + x(l + a)\} / 2^L$. This should be changed so that each word in the set has a different score based on its frequency. For example, click would have a value of 1, update would have a value of 1.2, and use would have a value of 1.4. This can also apply to the words contained in the set of words declaring a sense of urgency that affects the 'a' value and words conveying a sense of direction that affects the 'x' value as well. This change to the algorithm may require a change to the equation used in evaluation.

Another way to decrease false positive rate would be to increase the standard score used to classify an email as phishing to be more than 1. Word scores that were found in the phishing email corpus were as follows: 4733 scored 1.0, 8746 scored 1.5, and 870 scored 2.0. Word scores found in the legitimate email corpus were as follows: 2123 scored 1.0, 67 scored 1.5, and 9 scored 2.0. In the legitimate email corpus, 1.0 word scores mostly increased the false positive rate. One possible way to improve the algorithm is that if the score of the email is 1.0, then a user will be given a warning of that email instead of right throwing the email into a spam box.

Lastly, in this proposed algorithm, all POS are considered in order to check if the email has any keyword, and it proved that the new algorithm has merit of catching phishing emails, but it also has a serious drawback in terms of false positive. Alternately, a trade off can be an ideal option by finding what POS most frequently appears in phishing and legitimate emails instead of all POS. As emails are examined, the trade off point can be changed.

**CONCLUSION**

The paper expands a version of a text-based phishing detection algorithm. It is shown that though accounting for all POS rather than selective ones increases the phishing detection results. At the same time, false positive rate can slightly increase when more emphasis is put on the importance of detecting as many phishing emails as possible, as expected. It is expected that adding semantic component in addition to considering all POS will reduce the false positive rate while preserving the detection accuracy.

web sites. In *Proceedings of the 16th international conference on World Wide Web*, ACM, pp. 639-648.

**FIGURES AND TABLES**

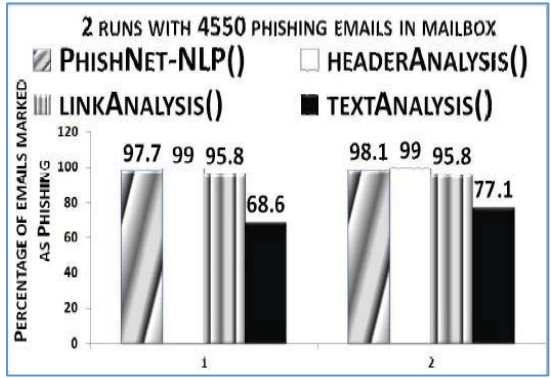

**Fig. 1 Results of Original PhishNet-NLP (Image from PhishNet-NLP). In the result of the first run which excluded the context score, the text analysis had 68.6% phishing detection rate.**

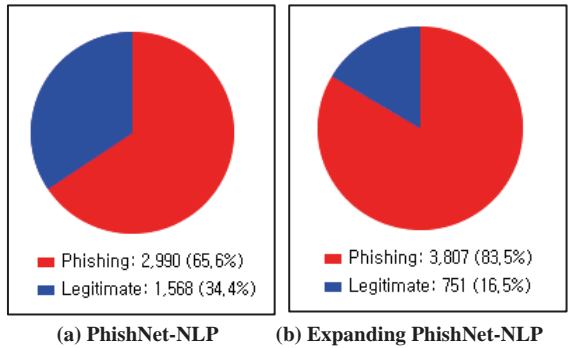

(a) PhishNet-NLP  (b) Expanding PhishNet-NLP

**Fig. 2 Phishing Corpus Results of Expanding PhishNet-NLP**

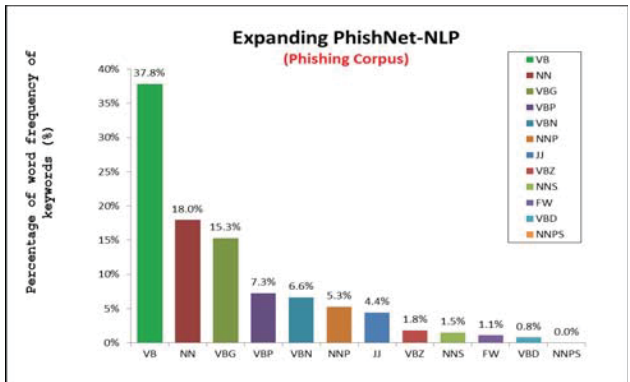

**Fig. 3 Phishing Corpus POS rankings in Expanding PhishNet-NLP**

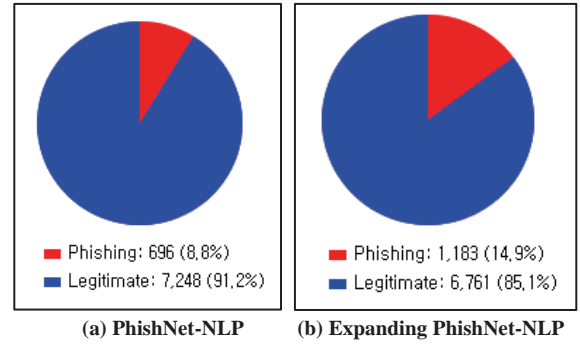

(a) PhishNet-NLP  (b) Expanding PhishNet-NLP

**Fig. 4 Legitimate Corpus Results of Expanding PhishNet-NLP**

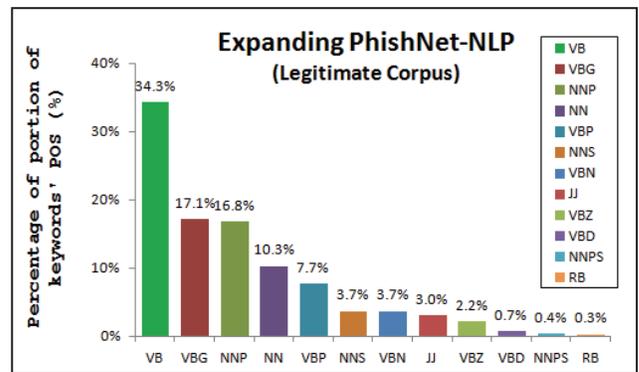

**Fig. 5 Legitimate Corpus POS rankings in Expanding PhishNet-NLP**

**Fig. 6 Stanford POS name abbreviations**